\documentclass[conference]{IEEEtran}
\IEEEoverridecommandlockouts
\usepackage{cite}
\usepackage{amsmath,amssymb,amsfonts}
\usepackage{algorithmic}
\usepackage{graphicx}

\usepackage{multirow}
\usepackage{textcomp}
\usepackage{booktabs}
\usepackage{subcaption}
\usepackage{xcolor}
\def\BibTeX{{\rm B\kern-.05em{\sc i\kern-.025em b}\kern-.08em
    T\kern-.1667em\lower.7ex\hbox{E}\kern-.125emX}}
\usepackage{hyperref}
\hypersetup{
    colorlinks=true,
    citecolor=blue,
    linkcolor=blue,
    filecolor=magenta,      
    urlcolor=cyan,
}

\begin{document}
\title{\vspace{0.7cm}
Differential Analysis of Multispectral Images for Terrain Identification%
\thanks{This paper was accepted at the 2026 IEEE/ASME International Conference on Advanced Intelligent Mechatronics (AIM 2026).}}
\author{
\IEEEauthorblockN{Omar Kashmar}
\IEEEauthorblockA{
\textit{University of Genoa}, Italy\\
omar.kashmar@edu.unige.it}
\and
\IEEEauthorblockN{Hmeandra Arya}
\IEEEauthorblockA{
\textit{Indian Institute of Technology Bombay}, Mumbai, India\\
arya@aero.iitb.ac.in}
\and
\IEEEauthorblockN{Fulvio Mastrogiovanni}
\IEEEauthorblockA{
\textit{University of Genoa}, Italy\\
fulvio.mastrogiovanni@unige.it}
}
\maketitle

\begin{abstract}
Reliable terrain understanding is a prerequisite for autonomous robot navigation.
Yet, the widespread RGB-based perception can fail under low illumination, shadows, and material ambiguities.
In this work we propose \textsf{DRIFT}, a lightweight multispectral framework that combines raw spectral bands and illumination-tolerant \emph{band-ratio} representations through a dual-stream residual architecture and a differential fusion branch. 
Band ratios attenuate multiplicative acquisition effects (illumination/sensor gains), while the differential fusion explicitly highlights discrepancies between absolute-band and ratio-derived cues, which improves the robustness to noisy or partially unreliable spectral measurements. 
In the paper 
(i) we evaluate \textsf{DRIFT} on a new \textit{oil-on-soil} multispectral dataset acquired using a MicaSense RedEdge-P camera mounted on an Unmanned Aerial Vehicle, and 
(ii) we provide an additional controlled study on \textit{water-on-grass} under varying illumination and thermal perturbations (hot/cold water) to analyze NIR-sensitive effects. 
\textsf{DRIFT} consistently improves over strong baselines, while remaining compatible with edge deployment. 
\end{abstract}


\noindent\textbf{Code Availability:} \url{https://tinyurl.com/4wzekcar}

\section{Introduction}
\label{sec:introduction}
A reliable identification of terrains is a prerequisite for a safe robot autonomy in a variety of outdoor scenarios.
A robot must anticipate what its wheels or feet will contact in the upcoming instants, often under challenging sensing conditions, such as low light, shadows, specularities, bad weather, or seasonal changes.
The majority of existing approaches rely on standard RGB imagery because it is ubiquitous and nowadays relatively inexpensive.
However, RGB is fundamentally limited when appearance changes are dominated by illumination rather than the properties of materials, or when different terrains exhibit near-identical color/texture cues, such as wet grass \textit{versus} muddy soil.
These limitations become critical in field robotics scenarios, in which sensing must remain robust and carried out in real time on embedded compute~\cite{c1, c24}.

\begin{figure}[t!]
\centering
\includegraphics[width=\linewidth]{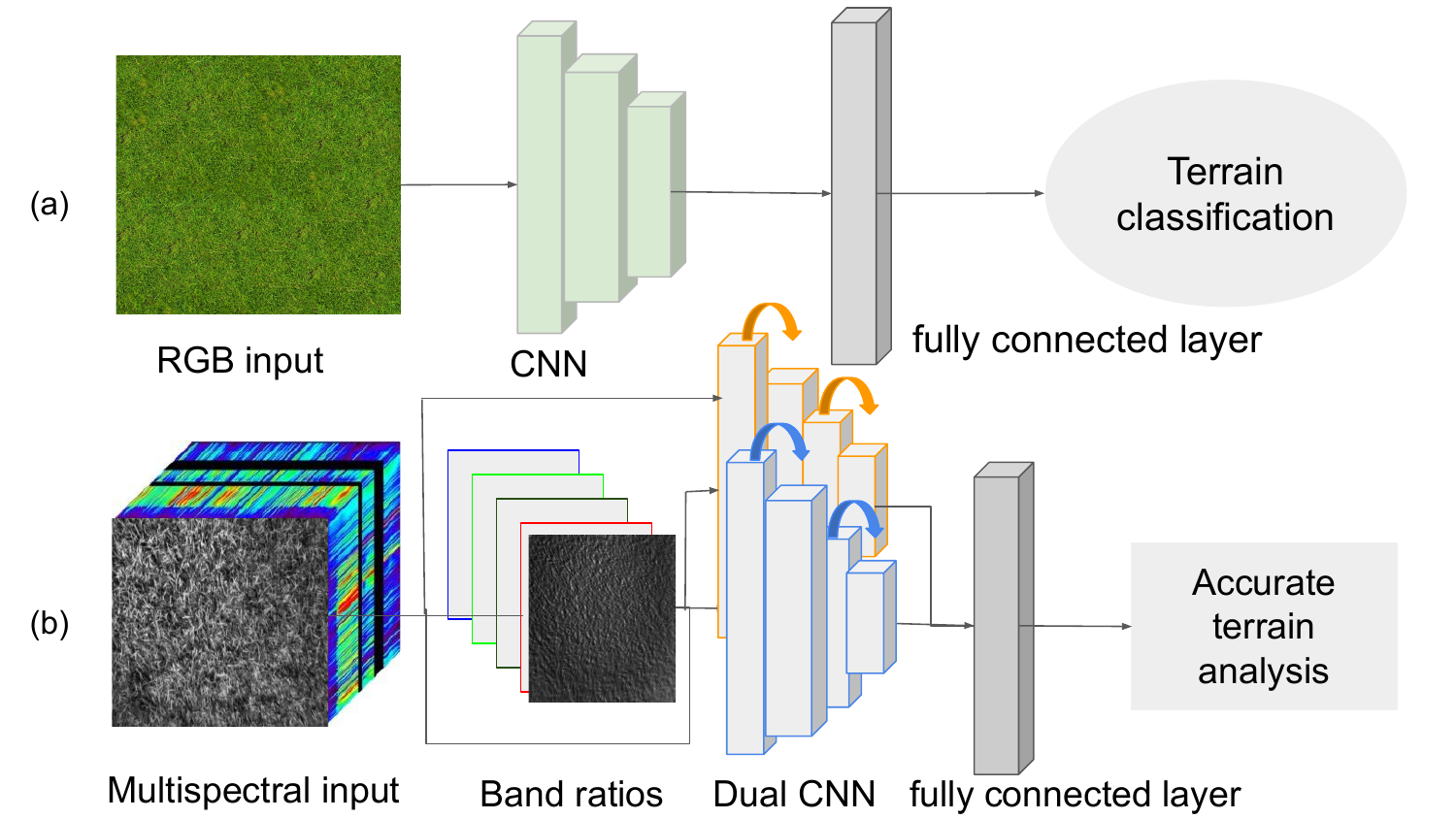}
\caption{
An accurate terrain analysis is required for outdoor robot navigation:
(a) RGB images may provide insufficient cues for robust terrain classification, for example, grass-covered dry, wet, or muddy terrain can appear similar to each other;
(b) multispectral images can enhance separability for materials such as oil and wet vegetation.
}
\label{fig:enter-label}
\end{figure}

Multispectral cameras may provide a practical middle ground between RGB and hyperspectral imaging.
They acquire a small number of informative bands, typically spanning visible and near-infrared, which are sensitive to physical and chemical properties of the surface. 
In particular, near-infrared responses can reveal information that is weak or ambiguous in RGB, such as vegetation condition and moisture content.
Specific spectral signatures can help distinguish slick or contaminated substrates, for example, oil films, from visually similar backgrounds~\cite{c11,c6}. 
Figure~\ref{fig:enter-label} illustrates a representative failure mode of RGB-based classification.
Multiple terrain states can look similar in the visible spectrum, while multispectral sensing preserves additional cues that can be exploited algorithmically.

Beyond cameras, alternative sensing modalities have been proposed for terrain and environment perception. 
LiDARs can be used to provide accurate geometry and high-resolution 3D mapping, and are largely invariant to illumination.
Nevertheless, they remain comparatively expensive and the computational cost of processing dense point clouds can be a limiting factor for edge deployment~\cite{c27, c1}.
Radar is robust at long range and under adverse weather conditions, but its lower spatial resolution can hinder fine-grained terrain discrimination~\cite{c1}. 
The Ground Penetrating Radar (GPR) can classify surface terrains for robots operating in harsh conditions and can complement semantic mapping frameworks.
However, it typically requires bulky and costly hardware configurations, which restricts its applicability on many mobile platforms~\cite{c2}.
As a result, vision remains attractive due to cost and richness of information, but requires improved robustness to the confounders that dominate in unstructured environments~\cite{c1}.

Recent learning-based approaches have explored cross-modal and self-supervised schemes to mitigate the fragility of pure RGB perception.
For example, the approach in~\cite{c7} uses visual input to infer terrain-related physical properties, such as stiffness and friction, with the aim to close the gap between simulation-trained policies and real-world deployment.
At the same time, the same approach highlights the sensitivity of visual data to environmental conditions.
In this context, multispectral perception is particularly promising because it encodes additional invariants linked to material composition and surface state.
\textit{Oil detection} is a salient example, as spectral signatures depend on oil properties and illumination, and near-infrared (NIR) bands have been shown to support oil spill detection in remote sensing settings~\cite{c11}.
Hyperspectral imaging can provide even richer spectral information by sampling narrow bands over a wide wavelength range~\cite{c11}, but its cost and complexity often make it impractical for robots. 

\begin{figure}[t!]
\centering
\includegraphics[width=0.96\linewidth]{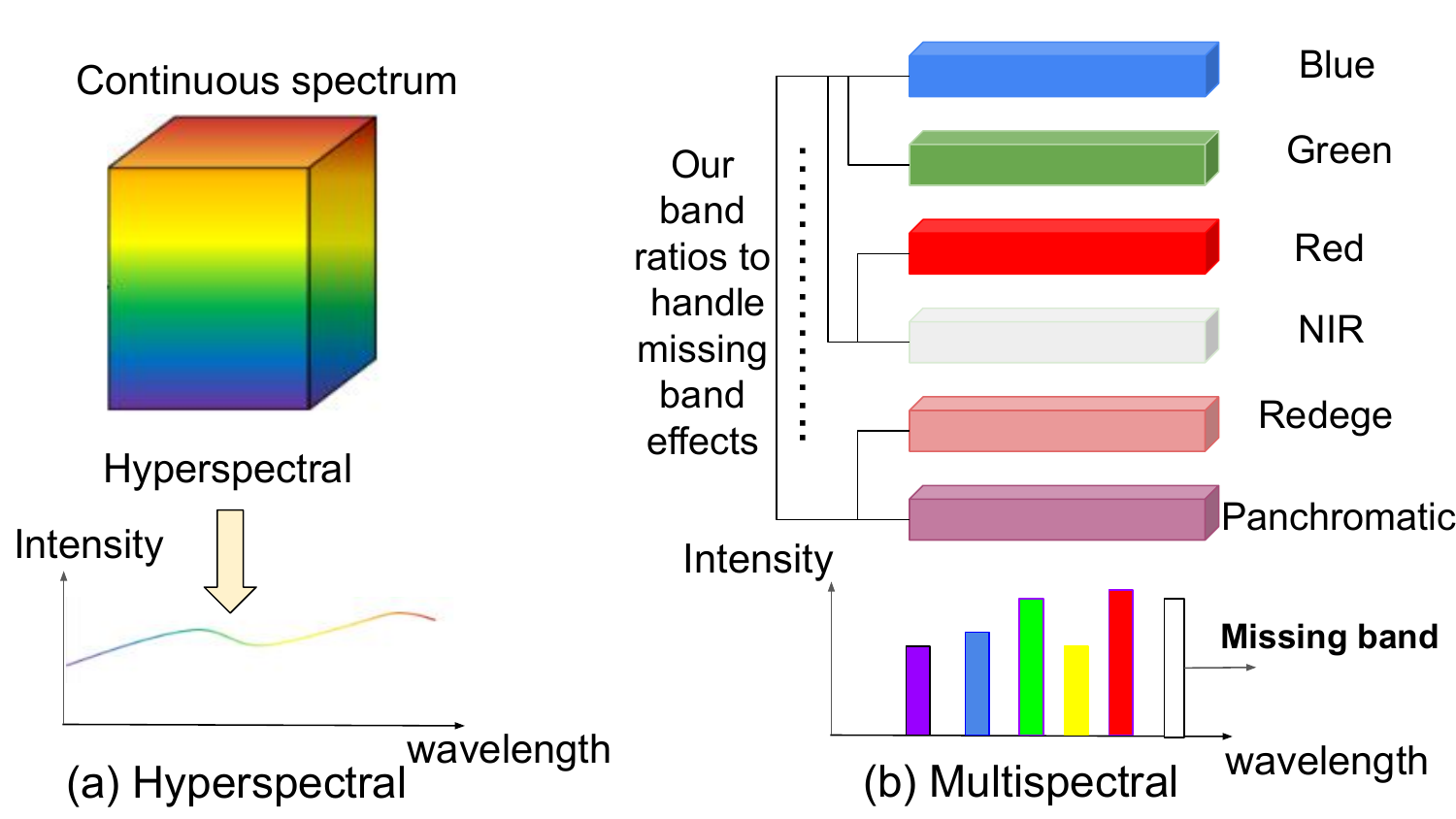}
\caption{
Comparison between traditional approaches and our approach:
(a) hyperspectral imaging provides densely sampled wavelengths over a continuous spectrum;
(b) in contrast, \textsf{DRIFT} uses multispectral band ratios to obtain robust, illumination-tolerant cues without the cost and complexity of hyperspectral cameras.
}
\label{fig:Drift}
\end{figure}

In this paper, we focus on 
(i) multispectral sensing as a deployable alternative, and
(ii) algorithmic mechanisms that extract robust cues without requiring expensive hyperspectral instrumentation.
In particular, this paper introduces \textsf{DRIFT}, which stands for \textit{Differential Ratio Integration For robust Terrain}, a framework that addresses robust terrain identification from multispectral imagery under realistic acquisition variability, including changes in illumination and environmental conditions.
Our core idea is to represent multispectral observations through \emph{band ratios}, and to learn complementary representations from both raw bands and ratio-derived cues as illustrated in Figure \ref{fig:Drift}. 
Band ratios are widely used in remote sensing because they can attenuate multiplicative illumination effects and emphasize material-dependent differences across wavelengths.
Building on this intuition, we introduce a dual-stream residual architecture that processes
(i) the original multispectral bands and band-ratio representations, and 
(ii) combines the resulting features through a differential fusion mechanism to improve separability under challenging conditions.

The main contributions of this work are:
(i) \textsf{DRIFT}, a multispectral differential framework for terrain identification as a learning-based approach that exploits multispectral band ratios and differential feature fusion to improve robustness over RGB-centric baselines;
(ii) an embedded-oriented design, whereas the proposed pipeline is designed to remain lightweight and suitable for deployment on edge compute commonly available on robot platforms;
(iii) an empirical evaluation under acquisition variability, for which we collect and analyze multispectral data under varying illumination and environmental conditions, including oil-on-soil scenes and water-on-grass scenarios, and we evaluate the impact of spectral cues (notably, NIR) on classification performance.

\section{Related Work}
\label{sec:related_work}

In the past few years, terrain understanding for autonomous robot has been approached through a wide range of sensing modalities and learning paradigms. 
While RGB cameras remain attractive for their low cost and high spatial resolution, their performance can degrade substantially under illumination changes, presence of shadows, weather variability, and material ambiguities, which produce similar visible appearance across different surface states~\cite{c1}. 
These limitations motivate either complementary sensors or representations that are less sensitive to illumination.

A significant body of work targets terrain recognition and traversability estimation using learning-based perception. 
Guan \textit{et al}.~\cite{a40} propose GA-Nav, a group-wise attention mechanism that performs efficient terrain segmentation for navigation in unstructured outdoor environments, and demonstrates strong results on established datasets. 
Beyond pure vision, alternative sensing channels and interaction-based methods have also been explored. 
For instance, terrain-related physical properties can be inferred from vision to support locomotion and navigation~\cite{c7}.
Other approaches leverage specialized sensing such as whisker-based systems combined with learning for terrain identification in unstructured environments~\cite{c15}, or acoustic cues for terrain classification~\cite{c19}. 
Self-supervised schemes can further reduce annotation effort by transferring supervision across data sources and learning terrain-related features directly from real-world imagery~\cite{c20}. 
However, many solutions still rely primarily on RGB cues, and robustness under challenging illumination remains a central open issue~\cite{c1}.

Multispectral imaging is widely adopted in remote sensing and in monitoring scenarios based on Unmanned Aerial Vehicles (UAVs), because it captures wavelength-dependent responses that correlate with material composition and surface state.
In the context of UAV-based sensing, Tsouros \textit{et al}.~\cite{a32} review precision agriculture applications and highlight the benefits of high-resolution data acquisition and timely decision-making, while also noting practical challenges related to standardized acquisition and processing pipelines. 
For environmental monitoring, multispectral imagery has been used for automated oil spill detection~\cite{c3} and operational response activities, including lessons learned from large-scale incidents such as the well-known Deepwater Horizon~\cite{c4_dwh}. 
Compared to hyperspectral imaging, which offers dense spectral sampling but typically involves higher sensor cost and processing complexity, multispectral setups provide a more practical trade-off for \textit{embedded} platforms. 
Hyperspectral classification methods can achieve high accuracy when rich spectral-spatial information is available~\cite{c21}.
However, multispectral approaches are often preferred when cost, payload, and compute constraints are strict. 
More generally, comparative analyses of multispectral \textit{versus} hyperspectral retrieval highlight fundamental trade-offs in spectral resolution, information content, and operational complexity~\cite{c5}.

A standard strategy in remote sensing is to compute spectral indices or band ratios to attenuate illumination effects and emphasize material-dependent differences across wavelengths. 
Band-ratio style representations have been used as effective inputs to learning pipelines in multispectral imagery analysis~\cite{a38}. 
In parallel, data fusion architectures, most notably dual-branch or dual-stream networks, have become a common mechanism to combine complementary representations.
Early multimodal segmentation architectures such as for example FuseNet~\cite{a33} inspired subsequent dual-branch designs in spectral-spatial classification~\cite{a39} and remote sensing tasks, including multispectral segmentation datasets and lightweight models targeting efficient deployment~\cite{a35, a37}.
Dual-branch spatial-spectral data fusion has also been explored for oil spill monitoring in high-dimensional spectral settings~\cite{a36}.

Building on these directions, \textsf{DRIFT} targets robot-relevant terrain identification under realistic acquisition variability, with particular emphasis on illumination robustness and deployability.
Differently from RGB-based terrain segmentation approaches~\cite{a40}, we explicitly leverage multispectral information to disambiguate visually similar terrain states. 
Compared to hyperspectral methods~\cite{c21}, we focus on a practical multispectral setup and derive discriminative cues through band ratios computed from a small number of bands. 
Finally, while prior dual-stream designs fuse modalities or representations~\cite{a33, a39}, we propose a differential fusion strategy that jointly exploits raw multispectral bands and band-ratio representations to improve robustness and sensitivity to oil/vegetation-related spectral signatures.
\vspace{-0.27cm}
\section{Methods}
\label{sec:methods}
\vspace{-0.17cm}

This Section describes the main conceptual tenets of \textsf{DRIFT}, which is based on 
(i) an illumination-tolerant ratio representation, and 
(ii) a dual-stream residual architecture with differential data fusion. 
The design is motivated by the need for robust classification under acquisition variability, that is,  illumination changes, shadows, and sensor gain differences, while remaining deployable on edge compute.

\subsection{Acquisition Model and Ratio Representation}
Let $X_i \in \mathbb{R}^{H \times W \times B}$ denote the $i$-th multispectral sample with $B$ spectral bands, where $H$ and $W$ are the image height and width (that is, spatial resolution in pixels), and let $X_i^{(b)} \in \mathbb{R}^{H \times W}$ be the image acquired in band $b \in \mathcal{B}=\{b_1,\dots,b_B\}$. 
A minimal and commonly used model for band-dependent illumination/sensor effects assumes that each observed band is approximately a scaled version of an underlying band-specific scene response \textit{plus} noise, that is,
\begin{equation}
X_i^{(b)} = g_i^{(b)}\, S_i^{(b)} + \eta_i^{(b)},
\label{eq:acq_model}
\end{equation}
where 
$g_i^{(b)}$ is a multiplicative factor accounting for illumination and sensor gain, 
$S_i^{(b)}$ captures the material-dependent reflectance-related signal, and 
$\eta_i^{(b)}$ is additive noise. 
While this model is simplified, it captures the key confounder that often dominates RGB-based terrain recognition in the field, that is, changes in intensity that do not correspond to changes in material.

To attenuate multiplicative effects, we represent multispectral data through \emph{band ratios}. 
For a predefined set of $K$ ordered band pairs $(p_k, q_k)$, with $p_k \neq q_k$, we define
\begin{equation}
R_i^{(k)} = \frac{X_i^{(b_{p_k})}}{X_i^{(b_{q_k})} + \epsilon}, 
\qquad k=1,\dots,K,
\label{eq:ratio_def}
\end{equation}
where $\epsilon>0$ avoids division by zero.
Under Equation~\eqref{eq:acq_model}, ratios reduce sensitivity to shared multiplicative factors and emphasize wavelength-dependent differences that are more tightly linked to material properties, that is, a standard practice in remote sensing and spectral analysis~\cite{c29, c30}. 
Stacking the ratios yields the ratio tensor
\begin{equation}
\mathcal{R}_i = \left[ R_i^{(1)}, \dots, R_i^{(K)} \right] \in \mathbb{R}^{H \times W \times K},
\label{eq:ratio_tensor}
\end{equation}
with $K \leq B(B-1)$ in the ordered-pair case. 
In practice, $K$ is selected to balance discriminative power and computational cost.

\subsection{Dual-stream Residual Encoders}
\textsf{DRIFT} learns complementary representations from
(i) the raw multispectral bands and 
(ii) the ratio tensor.
Specifically, we use two residual encoders:
\begin{equation}
F_X = f_{\theta}(X_i), 
\qquad
F_R = f_{\phi}(\mathcal{R}_i),
\label{eq:encoders}
\end{equation}
where $F_X, F_R \in \mathbb{R}^{H' \times W' \times D}$ are feature maps, and $f_{\theta}, f_{\phi}$ share the same backbone design (ResNet-style residual blocks), but do not share weights. 
The use of two streams allows the network to preserve
(a) absolute spectral content from $X_i$, and 
(b) illumination-tolerant relative cues from $\mathcal{R}_i$.

In \textsf{DRIFT}, we adopt a standard residual block formulation. 
Given an input feature map $H_{l-1} \in \mathbb{R}^{H' \times W' \times D}$, the $l$-th block computes
\begin{equation}
H_l = \mathrm{ReLU}\!\left(H_{l-1} + f_l(H_{l-1})\right),
\label{eq:resblock}
\end{equation}
where $f_l(\cdot)$ is a residual mapping implemented with two $3\times3$ convolutional layers, each followed by batch normalization and ReLU. This definition is used consistently in both streams.

\vspace{7.2pt}
\subsection{Differential Fusion and Classification Head}
To explicitly highlight discrepancies between absolute-band features and ratio-derived features, we compute a differential feature map, as
\begin{equation}
F_{\Delta} = g_{\psi}\!\left(\left|F_X - F_R\right|\right),
\label{eq:diff_feat}
\end{equation}
where $|\cdot|$ is the element-wise absolute difference, and $g_{\psi}$ is a lightweight refinement module, implemented as $\mathrm{Conv}^{3\times3}$ + $\mathrm{BN}$ + $\mathrm{ReLU}$, such that
\begin{equation}
F_{\Delta} = \mathrm{ReLU}\!\left(\mathrm{BN}\!\left(\mathrm{Conv}^{3\times3}\!\left(\left|F_X - F_R\right|\right)\right)\right).
\label{eq:diff_refine}
\end{equation}
We aggregate information from both streams and the differential branch. 
Let $\mathrm{GAP}(\cdot)$ denote global average pooling. 
We define the fused descriptor
\begin{equation}
z_i = \left[\mathrm{GAP}(F_X),\, \mathrm{GAP}(F_R),\, \mathrm{GAP}(F_{\Delta})\right] \in \mathbb{R}^{3D},
\label{eq:fused_descriptor}
\end{equation}
and obtain class probabilities with a small MLP and softmax, such that
\begin{equation}
p_{\omega}(y \mid X_i) = \mathrm{softmax}\!\left(\mathrm{MLP}_{\omega}(z_i)\right),
\label{eq:softmax}
\end{equation}
where $y \in \{1,\dots,C\}$, and $C$ is the number of terrain classes.

\subsection{Training Objective}
We train the full model end-to-end with the standard categorical cross-entropy loss:
\begin{equation}
\mathcal{L}_{\mathrm{CE}} = -\sum_{c=1}^{C} \mathbf{1}[y_i=c]\log\left(p_{\omega}(y_i=c \mid X_i)\right).
\label{eq:ce}
\end{equation}
In addition, to encourage discriminative structure in the differential representation, we optionally include a contrastive term computed on the pooled differential descriptor $\tilde z_i=\mathrm{GAP}(F_{\Delta})$, such that
\begin{equation}
\begin{aligned}
\mathcal{L}_{\mathrm{ctr}}
= \sum_{(i,j)} \Big(
&\|\tilde z_i-\tilde z_j\|_2^2\, \mathbf{1}[y_i=y_j] \\
&+ \max\!\big(0,\, m-\|\tilde z_i-\tilde z_j\|_2^2\big)\, \mathbf{1}[y_i \neq y_j]
\Big),
\end{aligned}
\label{eq:contrastive}
\end{equation}

where $m>0$ is a margin. 
The total objective is
\begin{equation}
\mathcal{L}_{\mathrm{total}} = \mathcal{L}_{\mathrm{CE}} + \lambda\,\mathcal{L}_{\mathrm{ctr}},
\label{eq:total}
\end{equation}
with $\lambda \geq 0$ controlling the contribution of the contrastive term. 
In our experiments we report results both with $\lambda=0$ (classification-only case) and with $\lambda>0$ to quantify the benefit of explicit differential-feature separation.

\noindent
\textit{Remark}.
The pipeline is intentionally lightweight.
Ratios are computed with simple per-pixel operations, and the differential branch uses only a small refinement module on top of two standard residual encoders. 
The ratio stream provides a representation that is less sensitive to multiplicative illumination variation, as in Equation~\eqref{eq:acq_model}, while the dual-stream differential fusion in Equation~\eqref{eq:diff_feat} promotes features that remain discriminative even when absolute intensity cues become unreliable.

\begin{figure}[t!]
\vspace{8pt}
\centering
\includegraphics[width=\linewidth]{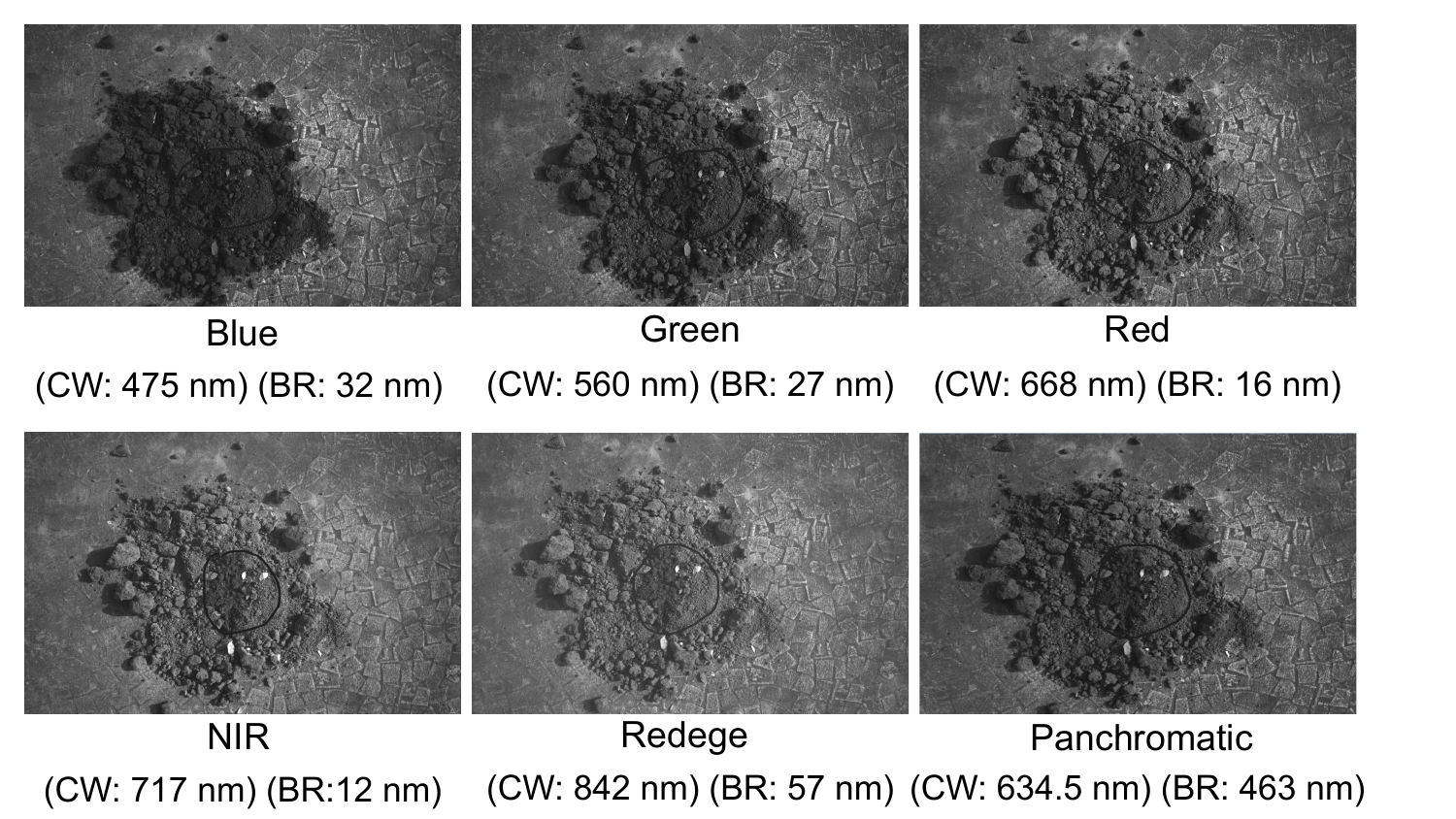}
\caption{
Example images showing the \textit{oil-on-soil} scenario, represented through six-band multispectral images.
`CW' is center wavelength, and `BR' is band range for each band.
}
\label{fig:oil}
\end{figure}

\begin{figure}[t!]
\centering
\includegraphics[width=\linewidth]{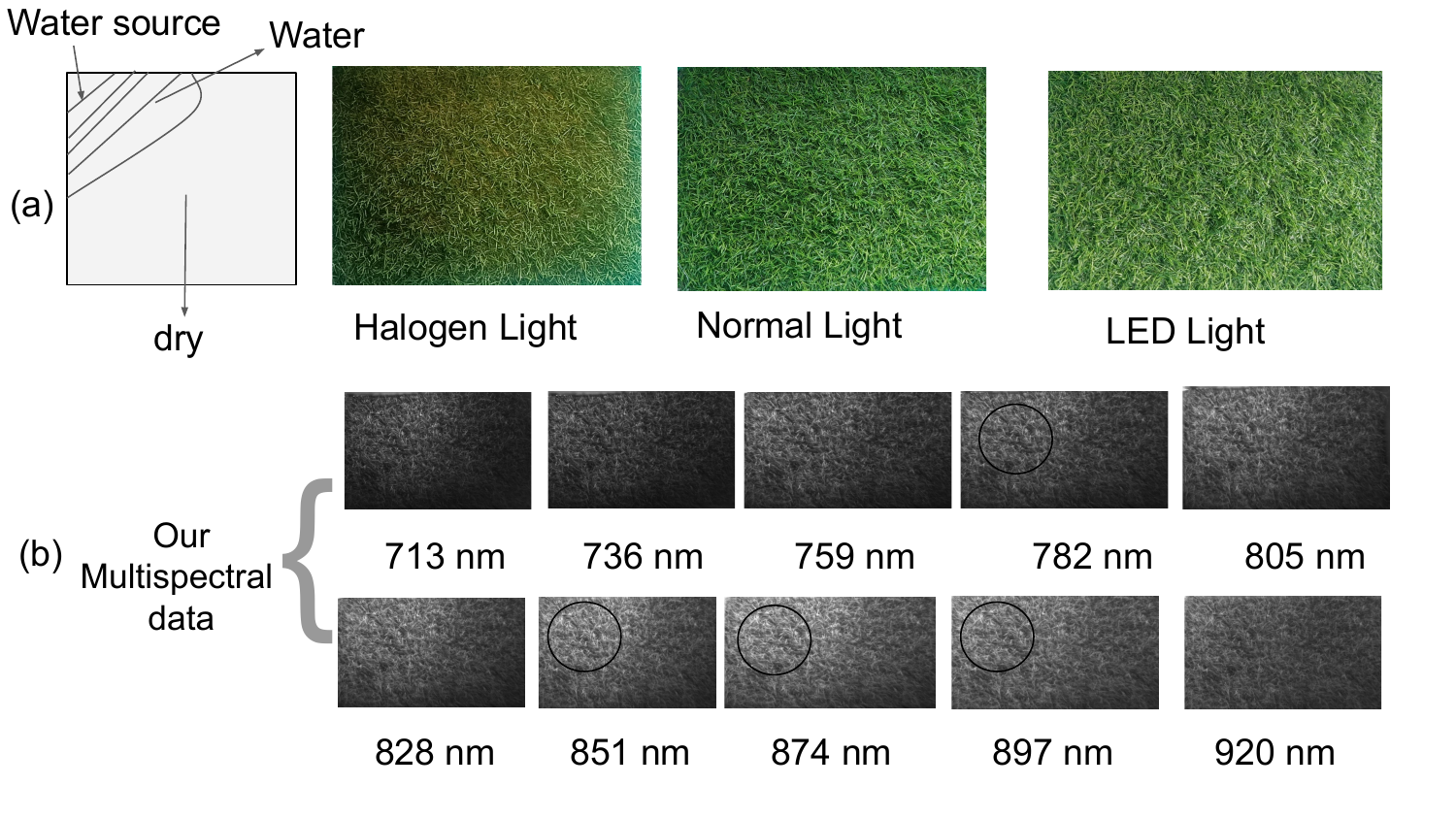}
\caption{
Use of multispectral data:
(a) the experiment setup and the RGB images for the grass terrain;
(b) multispectral images of grass terrain after the direct application of hot water under normal lighting condition.
}
\label{Figure: rgb}
\end{figure}

\begin{figure}[t!]
\centering
\begin{subfigure}[b]{0.48\columnwidth}
\centering
\includegraphics[width=\linewidth]{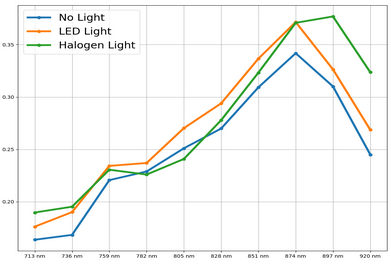}
\caption{\textit{Initial state}}
\label{fig:sub1}
\end{subfigure}
\hfill
\begin{subfigure}[b]{0.48\columnwidth}
\centering
\includegraphics[width=\linewidth]{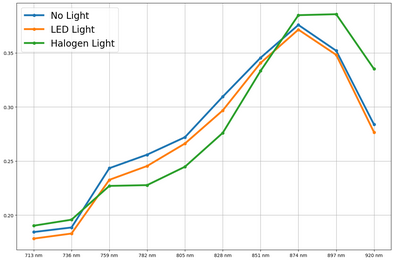}
\caption{\textit{at T1 (immediately)}}
\label{fig:sub3}
\end{subfigure}
\begin{subfigure}[b]{0.48\columnwidth}
\centering
\includegraphics[width=\linewidth]{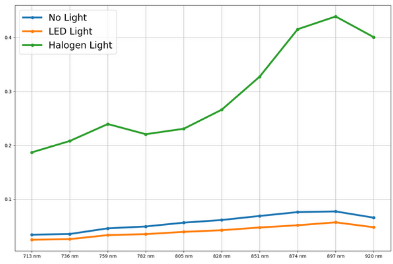}
\caption{\textit{at T2 (15 minutes later)}}
\label{fig:sub4}
\end{subfigure}
\hfill
\begin{subfigure}[b]{0.48\columnwidth}
\centering
\includegraphics[width=\linewidth]{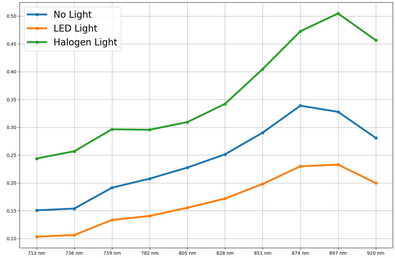}
\caption{\textit{after 1 hour}}
\label{fig:sub2}
\end{subfigure}
\caption{
Reflectance signals of grass terrain.}
\label{fig:reflect_2}
\end{figure}

\begin{figure}[t!]
\centering
\begin{subfigure}[b]{0.48\columnwidth}
\vspace{4pt}
\centering
\includegraphics[width=\linewidth]{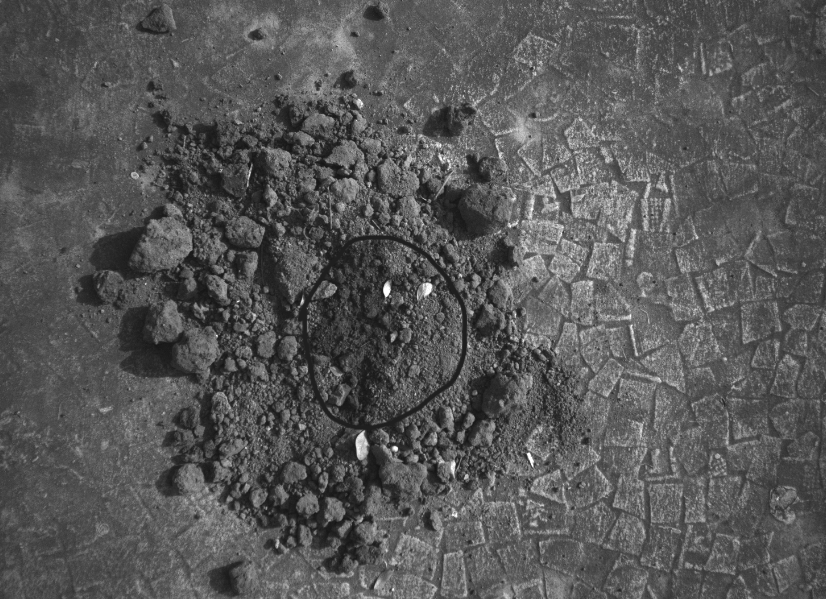}
\caption{Oil-affected}
\label{fig:conditions:sub1}
\end{subfigure}
\hfill
\begin{subfigure}[b]{0.467\columnwidth}
\centering
\includegraphics[width=\linewidth]{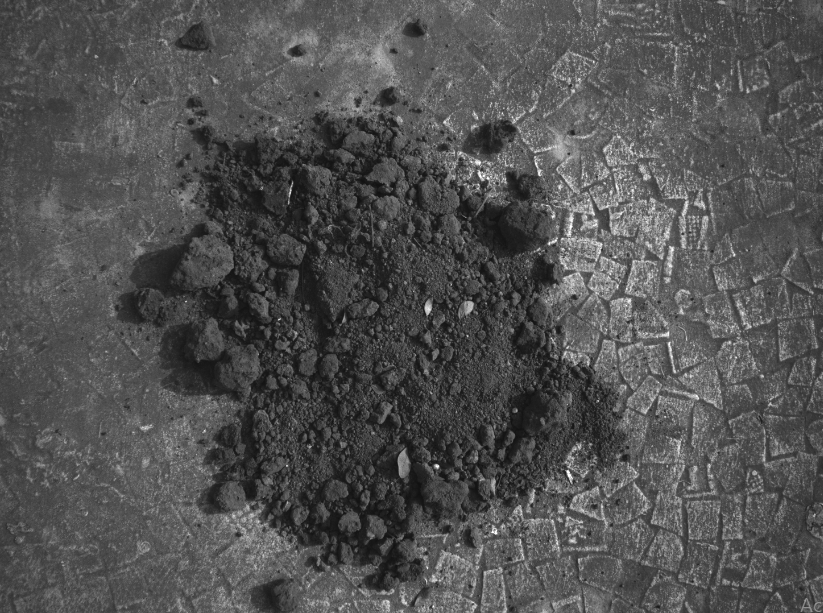}
\caption{Oil-free}
\label{fig:conditions:sub2}
\end{subfigure}
\caption{
Two experimental conditions:
(a) presence of oil;
(b) absence of oil.
}
\label{fig:exps1}
\end{figure}

\begin{figure}[t!]
\centering
\begin{subfigure}[b]{0.48\columnwidth}
\centering
\includegraphics[width=\linewidth]{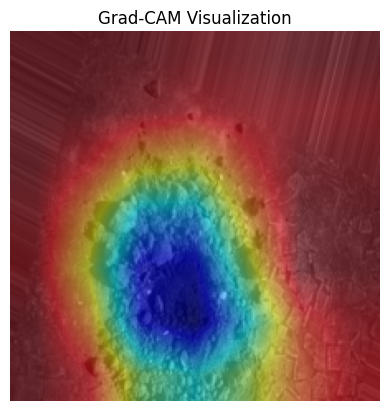}
\caption{Oil-affected}
\label{fig:grad_conditions:sub3}
\end{subfigure}
\hfill
\begin{subfigure}[b]{0.48\columnwidth}
\centering
\includegraphics[width=\linewidth]{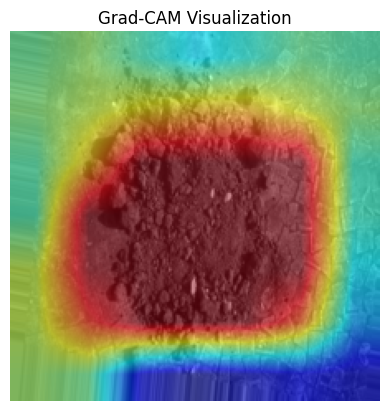}
\caption{Oil-free}
\label{fig:grad_conditions:sub4}
\end{subfigure}
\caption{
Grad-CAM visualizations for the two experimental conditions:
(a) presence of oil;
(b) absence of oil.
}
\label{fig:grad_conditions}
\end{figure}

\begin{figure}[t!]
\centering
\begin{subfigure}[b]{0.98\columnwidth}
\centering
\includegraphics[width=\linewidth]{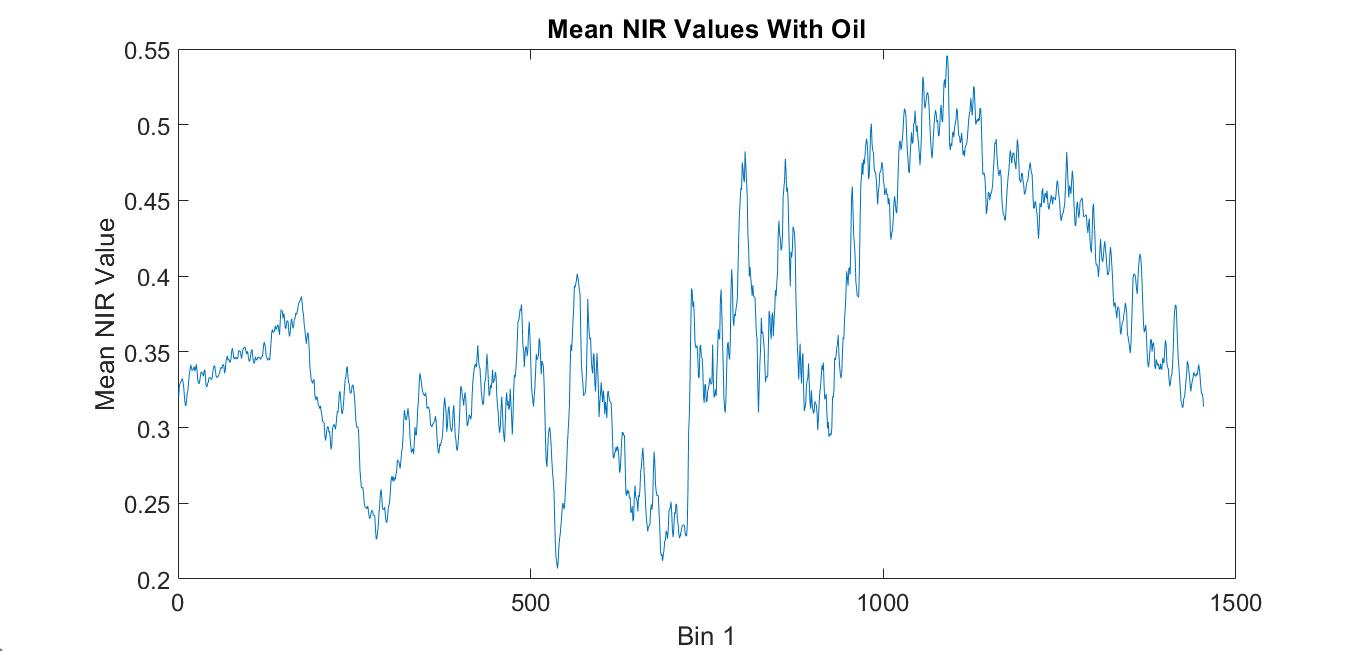}
\caption{Oil-affected}
\label{fig:NIRL:sub4}
\end{subfigure}
\\
\begin{subfigure}[b]{0.98\columnwidth}
\centering
\includegraphics[width=\linewidth]{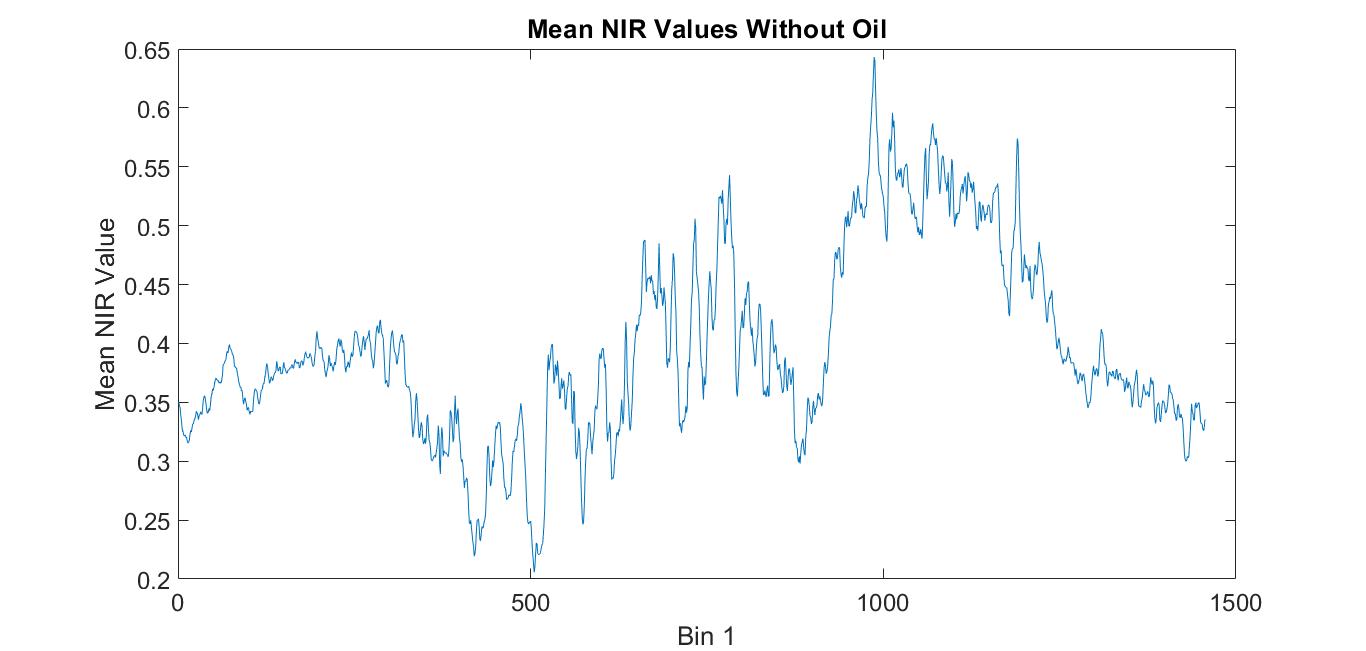}
\caption{Oil-free}
\label{fig:NIR:sub4}
\end{subfigure}
\caption{NIR statistics for the two experimental conditions:
(a) mean NIR pixel values for oil-affected samples;
(b) mean NIR pixel values for oil-free samples.
}
\label{fig:NIR}
\end{figure}

\vspace{0.4cm}
\section{Experimental Evaluation}
\label{sec:experiments}

We evaluate \textsf{DRIFT} on a multispectral \textit{oil-on-soil} dataset, and report an additional controlled \textit{water-on-grass} study to analyze the effect of illumination and thermal perturbations on NIR responses. 
All data were collected at the Indian Institute of Technology Bombay, Department of Aerospace Engineering. The experimental evaluation discussed here is designed to support three claims introduced in Sections~\ref{sec:introduction}--\ref{sec:methods}, namely 
(i) band-ratio cues improve robustness to acquisition variability,
(ii) dual-stream differential fusion provides complementary discriminative power over single-stream baselines, and 
(iii) the resulting model remains compatible with edge-oriented deployment.

\noindent
\textit{Datasets and acquisition protocol}. 
We collected an \textit{oil-on-soil} dataset using a UAV equipped with a MicaSense RedEdge-P multispectral camera with six spectral bands. 
Data were acquired in both static and flying modes.
In total, approximately $400$ images were captured while progressively adding oil to dry soil in multiple stages. 
After approximately $300$~ml of oil was poured in the central region, the contamination became visually more apparent in the red and NIR bands, while blue and green bands were comparatively less informative as illustrated in Figure \ref{fig:oil}. 
To reduce over-optimistic estimates caused by temporal correlation, for example, due to consecutive frames acquired during progressive oil addition, we adopted a \emph{sequence/stage}-level split rather than a frame-level split.
In addition, we performed controlled acquisitions in a \textit{water-on-grass} as illustrated in Figure \ref{Figure: rgb} scenario using a multi-band NIR camera (10 bands, $713$--$920$~nm) under different illumination sources, namely sunlight, LED, and halogen light. 
Hot water was applied to a designated region, and images were captured at three time instants, namely T1 (immediately after application), T2 (after $15$ minutes), and T3 (after $1$ hour).
This study is used as a supporting analysis to characterize how NIR responses vary under environmental perturbations relevant to outdoor robotic perception.

\noindent
\textit{Baselines and evaluation metrics}. 
For the \textit{oil-on-soil} classification task, we compare four models trained with identical optimization settings and the same train/validation/test split, as follows:
\begin{itemize}
\item
\textit{raw-only}: a single-stream encoder trained on the raw multi-spectral input $X_i$;
\item
\textit{ratio-only}: a single-stream encoder trained on the band-ratio tensor $\mathcal{R}_i$;
\item
\textit{concat-fusion}: a two-stream architecture in which raw-band and ratio-based features are fused by concatenation, without an explicit differential branch;
\item
\textsf{DRIFT}: the proposed two-stream architecture with differential fusion, as described in Section~\ref{sec:methods}.
\end{itemize}
For the six-band camera, the maximum number of ordered ratios is $B(B-1)=30$.
In our work, we use two complementary analyses: 
(i) single-ratio experiments, to assess the discriminative contribution of individual band ratios; and 
(ii) the full ratio-stream input used by \textsf{DRIFT}. 
We report classification accuracy and F1-score, and all results are averaged over three random seeds. 
\noindent
\textit{Main results on oil-on-soil classification}. 
Table~\ref{tab:main_results} summarizes quantitative results.
The progression from \textit{raw-only} to \textit{ratio-only}, then to \textit{concat-fusion}, and finally to \textsf{DRIFT}, is consistent with the conceptual motivation of our method.
Raw multi-spectral bands alone already provide a usable representation, but the ratio-based encoding improves discrimination, suggesting that band ratios help attenuate acquisition variability and highlight material-dependent spectral differences. 
A simple two-stream concatenation further improves performance, which indicates that raw and ratio-derived cues are complementary. 
However, the best results are obtained with \textsf{DRIFT}, whose explicit differential fusion yields the highest accuracy and F1-score, therefore supporting the hypothesis that modeling discrepancies between absolute-band and ratio-derived features is beneficial.

\begin{table}[t!]
\centering
\vspace{0.2cm}
\renewcommand{\arraystretch}{1.15}
\begin{tabular}{lcc}
\textbf{Method} & \textbf{Accuracy (\%)} & \textbf{F1-score} \\
\hline
\textit{raw-only} & 78.95 $\pm$ 0.11 & 0.714 $\pm$ 0.03 \\
\textit{ratio-only} & 85.70 $\pm$ 1.00 & 0.875 $\pm$ 0.01 \\
\textit{concat-fusion} & 87.72 $\pm$ 0.15 & 0.863 $\pm$ 0.10 \\
\textsf{DRIFT} & 94.50 $\pm$ 0.80 & 0.93 $\pm$ 0.02 \\
\end{tabular}
\caption{
Classification results for all internal baselines in the \textit{oil-on-soil} scenario.
Results are averaged over three random seeds.}
\label{tab:main_results}
\end{table}

\noindent
\textit{Effect of individual band ratios}. 
To better understand which spectral relationships are most informative, we trained classifiers using individual band-ratio images.
Table~\ref{tab:BandRatios} reports the resulting accuracies. 
It can be observed that ratios involving NIR tend to be among the most discriminative, which is consistent with the known sensitivity of NIR to surface composition and contamination.
In contrast, ratios between nearby visible bands often provide weaker discrimination. 
This observation is aligned with the physical intuition underlying \textsf{DRIFT}, and supports the use of ratio-derived cues as a robust complementary representation.

\begin{table}[t!]
\centering
\vspace{-0.58cm}
\renewcommand{\arraystretch}{1.05}
\begin{tabular}{llllll}
\multicolumn{6}{c}{\textbf{Single Band-Ratio Classification Accuracy (\%)}} \\ 
\midrule
\multicolumn{1}{l}{B1/B2} & \multicolumn{1}{l|}{98.4} & \multicolumn{1}{l}{B2/B1} & \multicolumn{1}{l|}{92.7} & \multicolumn{1}{l}{B1/B3} & 98.5 \\
\multicolumn{1}{l}{B2/B3} & \multicolumn{1}{l|}{85.0} & \multicolumn{1}{l}{B3/B2} & \multicolumn{1}{l|}{99.7} & \multicolumn{1}{l}{B1/B4} & 98.5 \\
\multicolumn{1}{l}{B3/B4} & \multicolumn{1}{l|}{98.4} & \multicolumn{1}{l}{B4/B3} & \multicolumn{1}{l|}{91.3} & \multicolumn{1}{l}{B1/B5} & 98.5 \\
\multicolumn{1}{l}{B4/B5} & \multicolumn{1}{l|}{98.4} & \multicolumn{1}{l}{B5/B4} & \multicolumn{1}{l|}{98.5} & \multicolumn{1}{l}{B1/B6} & 97.5 \\
\multicolumn{1}{l}{B5/B6} & \multicolumn{1}{l|}{96.8} & \multicolumn{1}{l}{B6/B5} & \multicolumn{1}{l|}{98.5} & \multicolumn{1}{l}{B2/B4} & 98.5 \\
\multicolumn{1}{l}{B2/B5} & \multicolumn{1}{l|}{98.5} & \multicolumn{1}{l}{B2/B6} & \multicolumn{1}{l|}{98.5} & \multicolumn{1}{l}{B3/B1} & 95.6 \\
\multicolumn{1}{l}{B3/B5} & \multicolumn{1}{l|}{95.6} & \multicolumn{1}{l}{B3/B6} & \multicolumn{1}{l|}{98.5} & \multicolumn{1}{l}{B4/B1} & 91.3 \\
\multicolumn{1}{l}{B4/B2} & \multicolumn{1}{l|}{94.2} & \multicolumn{1}{l}{B4/B6} & \multicolumn{1}{l|}{98.5} & \multicolumn{1}{l}{B5/B1} & 92.7 \\
\multicolumn{1}{l}{B6/B2} & \multicolumn{1}{l|}{99.9} & \multicolumn{1}{l}{B6/B3} & \multicolumn{1}{l|}{94.2} & \multicolumn{1}{l}{B6/B4} & 98.5
\end{tabular}
\caption{
Accuracy obtained by training on a single band-ratio image. 
Ratios involving NIR are generally more discriminative for \textit{oil-on-soil} separation.}
\label{tab:BandRatios}
\vspace{-1cm}
\end{table}

\noindent
\textit{Interpretability analysis}. 
To qualitatively verify that \textsf{DRIFT} relies on physically meaningful regions, we apply Gradient-weighted Class Activation Mapping (Grad-CAM) to the trained classifier.
Let $y^c$ denote the pre-softmax logit for class $c$, that is, \textit{oil}, and let $A^k \in \mathbb{R}^{H' \times W'}$ be the $k$-th activation map of the last convolutional layer, with $k = 1, \dots, K$. 
Grad-CAM computes a class-specific importance weight for each channel by globally averaging the gradients of $y^c$ with respect to $A^k$, that is
\begin{equation}
\alpha_k^c = \frac{1}{H'W'} \sum_{i=1}^{H'} \sum_{j=1}^{W'} \frac{\partial y^c}{\partial A_{ij}^k}.
\label{eq:gradcam_alpha_exp}
\end{equation}
The class activation map is then obtained as
\begin{equation}
L_{\mathrm{Grad\mbox{-}CAM}}^c = \mathrm{ReLU}\!\left(\sum_{k=1}^{K} \alpha_k^c A^k \right) \in \mathbb{R}^{H' \times W'}.
\label{eq:gradcam_map_exp}
\end{equation}
After bilinear upsampling to the input resolution, the resulting heatmaps indicate that the model focuses on the oil-affected regions as illustrated in Figure \ref{fig:grad_conditions}. The conditions of the experiment are illustrated in Figure \ref{fig:exps1}.
This observation suggests that the learned decision cues are spatially localized and physically plausible. 
In addition, the mean NIR pixel values exhibit a consistent shift between oil and non-oil samples, further supporting the importance of NIR for discrimination in this scenario as illustrated in Figure \ref{fig:NIR}.

\noindent
\textit{Computational aspects}. An 80/20 stratified split (random\_state=42) over seeds \{42,7,21\}. All models used Adam (lr=$10^{-4}$) + CE loss; DRIFT additionally used AdamW, label smoothing (0.1), LR scheduling, and early stopping (patience=5, 20 epochs).
We evaluated the computational footprint of all the considered models on an Intel i7-10700 CPU. 
The \textit{raw-only} and \textit{ratio-only} models each contain approximately $24.6$M parameters and require about $4.1$--$4.2$ GFLOPs per forward pass, whereas the two-stream variants roughly double both the parameter count and the computational cost.
In particular, \textsf{DRIFT} contains $48.67$M parameters and requires $8.34$ GFLOPs. 
On a single CPU core, its average inference latency is $180.77$ ms per image, while with 16 CPU threads the latency decreases to $71.21$ ms, corresponding to a throughput of $14.04$ images/s. 
These results indicate that, although the dual-stream design is naturally heavier than single-stream baselines, \textsf{DRIFT} remains compatible with edge-oriented CPU deployment, especially when moderate parallelism is available.

\noindent
\textit{Supporting study: water-on-grass}. 
As a supporting analysis, we considered the \textit{water-on-grass} scenario under varying illumination and thermal perturbations.
Across both regular and LED illumination, the reflectance peak around $897$~nm remains stable, while the localized thermal perturbation induced by hot water produces a decrease that becomes more evident over time as illustrated in Figure \ref{fig:reflect_2}. 
Although this study is not used as a benchmark classification task, it provides qualitative evidence that NIR-sensitive responses vary in a structured way with environmental changes, thereby motivating the use of multi-band cues and ratio-based representations for outdoor robotic perception.

\noindent
\textit{Discussion}. 
The experimental results support three main observations. 
\textit{First}, band ratios are effective because, under the simplified acquisition model of Section~\ref{sec:methods}, they attenuate multiplicative illumination and gain effects while emphasizing wavelength-dependent differences associated with the underlying material. 
This is empirically supported by the strong performance of several NIR-involving ratios in Table~\ref{tab:BandRatios}. 
\textit{Second}, differential fusion provides additional value beyond ratios alone. 
While ratio-based encodings are robust, they may suppress useful absolute spectral cues. 
\textsf{DRIFT} achieves substantially better performance than all internal baselines \textit{because} it jointly processes raw bands and ratio-derived features, and because it explicitly models their discrepancy.
\textit{Third}, the present evaluation has limitations. The \textit{oil-on-soil} dataset remains relatively small and scenario-specific, and the current results are limited to one primary sensor configuration and one contamination setup. 
Broader generalization across soil types, oil types, illumination geometries, and sensors should therefore be considered in future work.
A natural next step is to complement the current evaluation with controlled robustness experiments under synthetic gain perturbations, additive noise, and missing-band masking.

\section{Conclusion}
\label{sec:conclusion}

In this paper, we present \textsf{DRIFT}, a lightweight multispectral framework for terrain surface classification that combines raw spectral bands with illumination-tolerant \emph{band-ratio} representations using a dual-stream residual architecture and a differential fusion branch. 
The key idea of our work is to exploit complementary cues: on the one hand, band ratios attenuate multiplicative acquisition effects, such as illumination and sensor gain variations; on the other hand, raw-band features preserve absolute spectral information that can be lost by normalization. 

Experiments on an \textit{oil-on-soil} dataset acquired with a MicaSense RedEdge-P camera indicate that
(i) multispectral ratios, especially those involving NIR, provide strong discriminative power for oil contamination, and that 
(ii) differential fusion improves robustness relative to single-stream baselines. 
Grad-CAM visualizations further suggest that the model focuses on physically meaningful regions around the contaminated area, supporting the interpretability of the learned decision cues. 
In addition, a controlled \textit{water-on-grass} study under varying illumination and thermal perturbations provides qualitative evidence that NIR responses are sensitive to environmental changes, motivating the use of multi-band cues and ratio-based normalization for outdoor robot perception.

This work has limitations.
The current evaluation is based on a single primary acquisition setting and a limited set of surface conditions, whereas temporal correlation in progressive contamination sequences can inflate performance if not handled with sequence-wise splits. 
Future work will therefore
(i) expand the dataset across different soils, oil types, illumination geometries, and sensors; 
(ii) include explicit robustness tests.

\bibliographystyle{IEEEtran}  
\bibliography{refrence}  

\end{document}